\documentclass[11pt,a4paper]{article}
\usepackage[hyperref]{naaclhlt2018}
\usepackage{times}
\usepackage{latexsym}
\usepackage[utf8]{inputenc}
\usepackage{caption}
\usepackage{multirow}
\usepackage{graphicx}
\usepackage{graphicx}
\usepackage[normalem]{ulem}
\useunder{\uline}{\ul}{}

\usepackage{url}

\aclfinalcopy % Uncomment this line for all SemEval submissions

\title{Team Neuro at SemEval-2020 Task 8: Multi-Modal Fine Grain Emotion Classification of Memes using Multitask Learning }

\author{Sourya Dipta Das \\
  Jadavpur University \\
  Kolkata, India \\
  {\tt dipta.juetce@gmail.com} \\\And
  Soumil Mandal \\
  The University of Texas at Dallas\\
  Dallas, USA \\
  {\tt soumil.mandal@gmail.com} \\}

\date{}

\begin{document}
\maketitle
\begin{abstract}
In this article, we describe the system that we used for the memotion analysis challenge, which is Task 8 of SemEval-2020. This challenge had three subtasks where affect based sentiment classification of the memes was required along with intensities. The system we proposed combines the three tasks into a single one by representing it as  multi-label  hierarchical  classification problem.Here,Multi-Task learning or Joint learning Procedure is used to train our model.We have used dual channels to extract text and image based features from  separate Deep Neural Network Backbone and aggregate them to create task specific features. These task specific aggregated feature vectors ware then passed on to smaller networks with dense layers, each one assigned for predicting one type of fine grain sentiment label. Our Proposed method show the superiority of this system in few tasks to other best models from the challenge.
\end{abstract}

\section{Introduction}

With the rise in popularity of social media, the usage of internet meme has been increasing as well to convey messages or reactions in a unique way. Most of the automatic sentiment classifiers in the past has been designed either for text or images, but for memes, we see that there's both the textual and visual component, which introduces the challenge of multi-modal aspects. Majority of these memes are used for performative acts, which generally have some kind of a sentiment for the ongoing social discourse. A lot of times, malicious users post disturbing or hate memes, which are often flagged my moderators. A lot of companies hire people to do this job, but with the rapidly growing volume, it becomes an important task to detect this in an automated manner which is scalable. It is much more challenging than simple text or image classification as the intent is described by the combined effect of both, i.e. visual cue and language understanding. The two main challenges are identifying or extracting the text from the image, and secondly, connecting the text with the image to get the sentiment, which often required domain knowledge. For example, a meme which has the marvel character Captain America in it, often employs the domain knowledge that he is honorable and responsible to generate humour combined with the text.

The organizers of SemEval 20 Task 8 created a dataset containing of x internet memes and proposed three tasks. Task A - sentiment classification, where the goal is to classify it to positive, negative or neutral. Task B - humor classification, classify memes to humor categories like sarcastic, humorous, offensive and motivation. Task C - scales of semantic class, quantify extent of humors into scales, not, slightly. We propose a multi-task learning system which uses combined feature from ResNet based CNN Model for Image feature extractor block \& recurrent DNN Model consists of stacked layers of bidirectional LSTM and GRUs with contextual attention as text feature extractor to learn all the three tasks at once. Our model gets best Macro-F1 scores of 0.3488, 0.5112, 0.3240 and corresponding Micro - F1 Score of 0.5021, 0.3998 on Tasks A, B and C respectively.

\begin{figure}
  \centering
  \includegraphics[height=7.5cm,trim=1 1 0 0,clip]{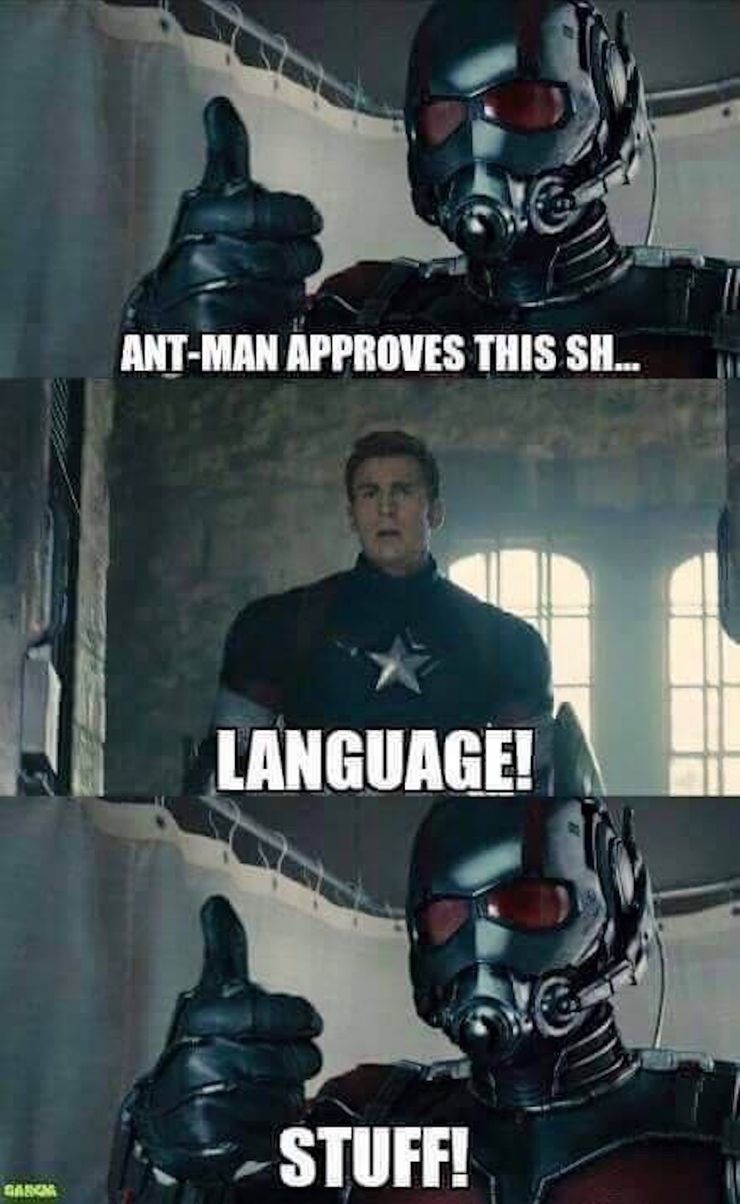}
  \caption{Sample Meme Image which describes the problem regarding Prior Domain Knowledge of the marvel character Captain America}
\label{fig21}
\end{figure}

\section{Related Work}
A transfer learning approach along with convolutional networks was employed by \citet{islam2016visual} for binary sentiment classification on images. A weakly supervised coupled CNN based method was proposed by \citet{yang2018weakly} to leverage localized information of images to predicts its sentiment. In \citet{sivakumar2018rosetta}, the authors proposed a model which uses R-CNN and fully convolutional model with CTC loss for understanding text from images and videos. \citet{gaspar2019multimodal} took the challenge of multimodal sentiment analysis where the data is in the form of image + related text. They created three separate classifiers, which were text classifier, image classifier and an image content analyzer. The predictions of all the three classifiers were combined in a weighted voted manner to give the final prediction. \citet{xu2019sentiment} developed a novel hierarchical deep fusion model for learning cross-modal relations between images and text, which then can be utilized for more effective sentiment analysis. \citet{xi2020multimodal} developed models using multi-head attention mechanism for sentiment classification on multimodal datasets like \citet{perez2013utterance} and \citet{zadeh2016mosi} where there were both visual and audio features. 

\section{Dataset Details}
We are using memotion dataset 7k\cite{chhavi2020memotion}\footnote{\url{https://www.kaggle.com/williamscott701/memotion-dataset-7k}} for out experiments. As there were three tasks for this challenge, the dataset was tagged in three ways. Coarse level distribution of the train and validation data post cleaning was 6601 and 914 respectively of each type of class, i.e. overall sentiment, humour, sarcasm, offensive and motivational. The fine-grain distribution of training and validation is shown in Table~\ref{table1}.

\begin{table}[h]
\centering
\captionsetup{justification=centering}
\resizebox{\columnwidth}{!}{%
\begin{tabular}{|c|c|l|c|l|}
\hline
\textbf{Overall} & \multicolumn{2}{c|}{\begin{tabular}[c]{@{}c@{}}positive : 3928\\ negative : 2089\\ neutral : 584\end{tabular}} & \multicolumn{2}{c|}{\begin{tabular}[c]{@{}c@{}}positive : 564\\ negative : 71\\ neutral : 279\end{tabular}} \\ \hline
\textbf{Humour} & \multicolumn{2}{c|}{\begin{tabular}[c]{@{}c@{}}not funny : 1577\\ funny : 2317\\ very funny : 2096\\ hilarious : 611\end{tabular}} & \multicolumn{2}{c|}{\begin{tabular}[c]{@{}c@{}}not funny : 210\\ funny : 315\\ very funny : 310\\ hilarious : 79\end{tabular}} \\ \hline
\textbf{Sarcasm} & \multicolumn{2}{c|}{\begin{tabular}[c]{@{}c@{}}no sarcasm : 1470\\ general : 3301\\ very twisted : 366\\ twisted meaning : 1464\end{tabular}} & \multicolumn{2}{c|}{\begin{tabular}[c]{@{}c@{}}no sarcasm : 233\\ general : 444\\ very twisted : 42\\ twisted meaning : 195\end{tabular}} \\ \hline
\textbf{Offensive} & \multicolumn{2}{c|}{\begin{tabular}[c]{@{}c@{}}not offensive : 2580\\ slightly offensive : 2436\\ very offensive : 1378\\ hateful offensive : 207\end{tabular}} & \multicolumn{2}{c|}{\begin{tabular}[c]{@{}c@{}}not offensive : 369\\ slightly offensive : 323\\ very offensive : 198\\ hateful offensive : 24\end{tabular}} \\ \hline
\multicolumn{1}{|l|}{\textbf{Motivational}} & \multicolumn{2}{c|}{\begin{tabular}[c]{@{}c@{}}not motivational : 4280\\ motivational : 2321\end{tabular}} & \multicolumn{2}{c|}{\begin{tabular}[c]{@{}c@{}}not motivational : 589\\ motivational : 325\end{tabular}} \\ \hline
\end{tabular}%
}
\caption{Class distribution in train and validation sets.}
\label{table1}
\end{table}

\begin{figure*}[h]
\centering
\includegraphics[scale=0.5]{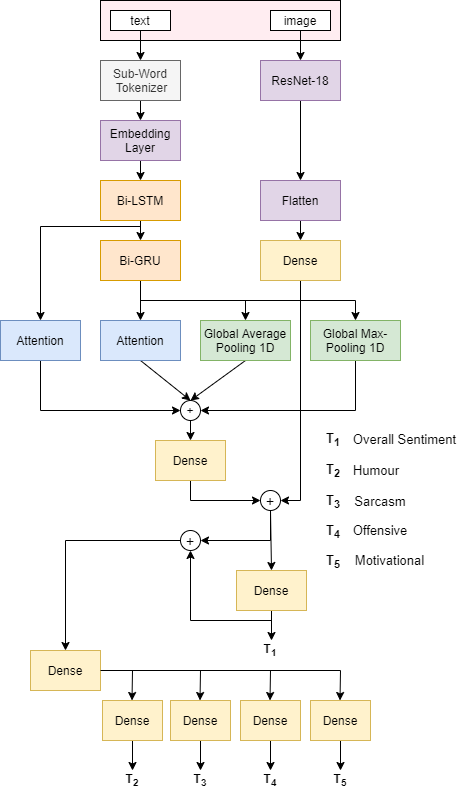}
\caption{Architecture overview.}
\label{fig1}
\end{figure*}

\begin{figure*}
  \centering
  \includegraphics[height=6.5cm, width=\textwidth,trim=1 1 0 0,clip]{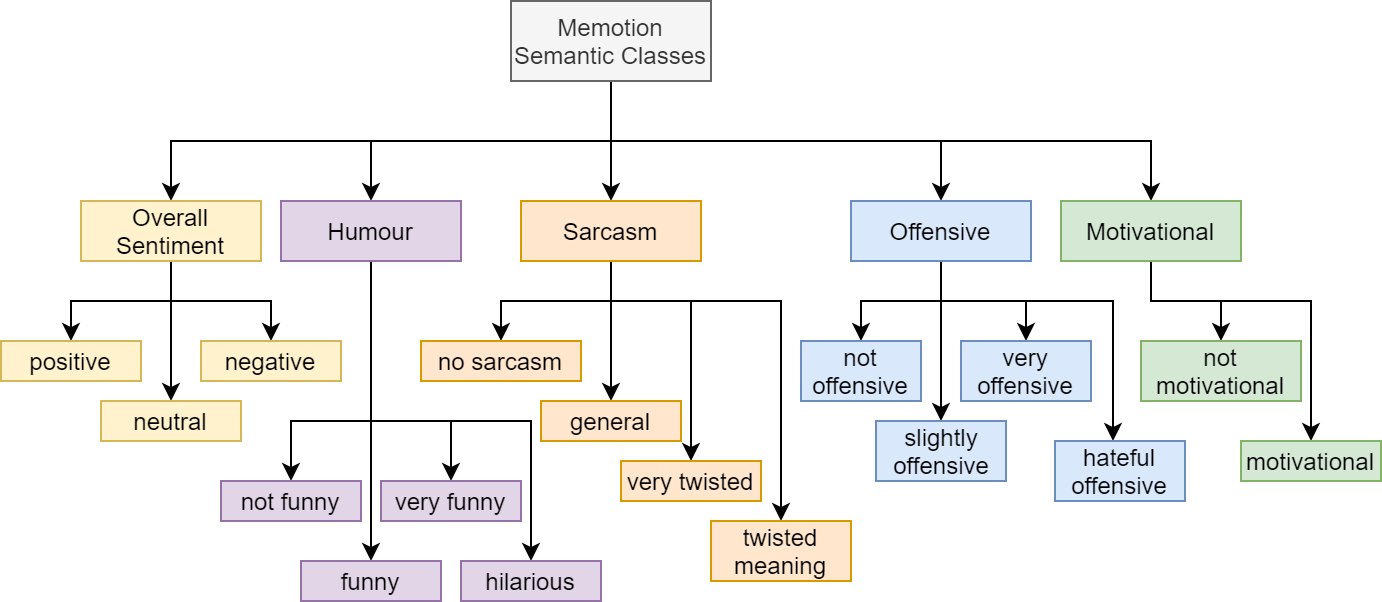}
  \caption{Semantic class hierarchy.}
\label{fig2}
\end{figure*}

\section{Proposed Method}
Our proposed method is based on multitask learning by jointly learning overall sentiment and  humor with Scales of corresponding semantic classes. In our system, we have represented this problem as multi-label hierarchical classification problem. Here, we create five main semantic class for each meme image and text pair, which are are 1) overall sentiment 2) motivational 3) offensive 4) sarcasm 5) humour. For each primary semantic class, more fine grained classes are assigned to them, which are used to quantify the extent to which a particular semantic class is being expressed in that meme image. In this way, we can gather the three tasks into one single problem. The hierarchical tree of classes is illustrated in the Figure~\ref{fig2} to depict the relationship between fine grain classes and their parent main semantic class. 
For testing phase, we created a separate model for each semantic class with same deep neural architecture but different learned parameters. In that way, each distinct model can expertise in learning semantic feature of corresponding semantic class assigned for that model. During the training phase, meme image and corresponding cleaned meme text in pairs is fed into the joint model. The trained model produces prediction for all five semantic classes as output. At high level, internal modules of the joint model is shown in the Figure~\ref{fig1}. Our joint model architecture consists of three main parts, which are 1) image feature extraction 2) text feature extraction
and 3) feature aggregation. Details on each part of the model is described in the subsequent subsections.

\subsection{Image Feature Extraction}
In the image feature extraction module, semantic class specific feature is extracted from the resized meme input image. In the preprocessing step, input meme image is resized into (224x224) sized image . After that, it was fed into a deep convolutional neural network from which the output activation map was flattened to get the image feature vector. Then, it was passed into the dense layer to get the final compressed image feature vector for aggregation. The main purpose of this dense layer is dimensionality reduction of raw flattened image feature from the CNN. No spatial augmentation was performed on the input image deliberately as it can confuse the CNN layer more in discriminative feature generation.

\subsection{Text Feature Extraction}
In this module, discriminative feature from text is extracted and fed into the later aggregation module. In pre-processing step, input texts are lower cased and stop word, web URLs were removed from the given input and cleaned text was extracted from meme image. Then, preprocessed text was fed into subword-tokenizer to get initial subword token with it's corresponding subword embedding. To get subword embedding vector for each token, we have fitted SentencePiece Processor~\cite{kudo2018sentencepiece}\footnote{\url{https://github.com/google/sentencepiece}} with all of our preprocessed clean text data. Then, these subword tokenized vector for each text is fed into a another trainable embedding layer to generate more specific embedding for each subword token. Finally, embedding vector for each text sequence is passed though stacked layers of bidirectional LSTM and GRU with multihead contextual attention layer followed by global average and max pooling layers to get final text feature vector from final fully connected dense layer. Here, the dense layer is used for dimensionality reduction of raw text feature from the stacked deep recurrent neural networks.

\begin{table*}[!htb]
\centering
\captionsetup{justification=centering}
\resizebox{\textwidth}{!}{%
\begin{tabular}{|c|c|c|c|c|}
\hline
Method & \begin{tabular}[c]{@{}c@{}}Macro-F1 Score\\  from \\ Our Method\end{tabular} & \begin{tabular}[c]{@{}c@{}}Best Macro-F1 Score \\ from\\ Leaderboard\end{tabular} & \begin{tabular}[c]{@{}c@{}}Micro - F1 Score \\ from \\ Our Method\end{tabular} & \begin{tabular}[c]{@{}c@{}}Micro - F1 Score \\ from \\ Leaderboard\end{tabular} \\ \hline
SE-ResNet18 & \textit{0.3488} & \multirow{4}{*}{\textbf{0.3546}} & \textit{0.5021} & \multirow{4}{*}{0.4872} \\ \cline{1-2} \cline{4-4}
ResNet18 & 0.3474 &  & \textbf{0.5202} &  \\ \cline{1-2} \cline{4-4}
ResNet34 & 0.3363 &  & 0.5186 &  \\ \cline{1-2} \cline{4-4}
VGG & 0.3297 &  & 0.4968 &  \\ \hline
\end{tabular}%
}
\caption{Task A - Comparison between best model from leaderboard and our proposed model with different backbone for visual feature extraction.}
\label{res1}
\end{table*}

\begin{table*}[!htb]
\centering
\captionsetup{justification=centering}
\resizebox{\textwidth}{!}{%
\begin{tabular}{|c|c|c|c|c|}
\hline
Method &
  \begin{tabular}[c]{@{}c@{}}Macro-F1 Score\\  from \\ Our Method\end{tabular} &
  \begin{tabular}[c]{@{}c@{}}Best Macro-F1 Score \\ from\\ Leaderboard\end{tabular} &
  \begin{tabular}[c]{@{}c@{}}Micro - F1 Score \\ from \\ Our Method\end{tabular} &
  \begin{tabular}[c]{@{}c@{}}Micro - F1 Score \\ from \\ Leaderboard\end{tabular} \\ \hline
SE-ResNet18 & \textit{0.4975} & \multirow{4}{*}{\textbf{0.5183}} & \textit{0.6017} & \multirow{4}{*}{0.6144} \\ \cline{1-2} \cline{4-4}
ResNet18    & 0.4686          &                                  & \textbf{0.6685} &                         \\ \cline{1-2} \cline{4-4}
ResNet34    & 0.5112          &                                  & 0.5988          &                         \\ \cline{1-2} \cline{4-4}
VGG         & 0.5077          &                                  & 0.6071          &                         \\ \hline
\end{tabular}%
}
\caption{Task B - Comparison between best model from leaderboard and our proposed model with different backbone for visual feature extraction.}
\label{res2}
\end{table*}
\begin{table*}[!htb]
\centering
\captionsetup{justification=centering}
\begin{tabular}{|c|c|c|c|c|}
\hline
Method & \begin{tabular}[c]{@{}c@{}}Macro-F1 Score\\  from \\ Our Method\end{tabular} & \begin{tabular}[c]{@{}c@{}}Best Macro-F1 Score \\ from\\ Leaderboard\end{tabular} & \begin{tabular}[c]{@{}c@{}}Micro - F1 Score \\ from \\ Our Method\end{tabular} & \begin{tabular}[c]{@{}c@{}}Micro - F1 Score \\ from \\ Leaderboard\end{tabular} \\ \hline
SE-ResNet18 & \textit{0.3125} & \multirow{4}{*}{0.3224} & \textit{0.3924} & \multirow{4}{*}{0.3779} \\ \cline{1-2} \cline{4-4}
ResNet18 & 0.25066 &  & \textbf{0.4402} &  \\ \cline{1-2} \cline{4-4}
ResNet34 & \textbf{0.3240} &  & 0.3998 &  \\ \cline{1-2} \cline{4-4}
VGG & 0.3192 &  & 0.4041 &  \\ \hline
\end{tabular}
\caption{Task C - Comparison between best model from leaderboard and our proposed model with different backbone for visual feature extraction.}
\label{res3}
\end{table*}
% Please add the following required packages to your document preamble:
% \usepackage{multirow}
\begin{table*}[!htb]
\centering
\captionsetup{justification=centering}
\begin{tabular}{|c|c|c|c|c|c|c|c|c|}
\hline
\multirow{2}{*}{Method} & \multicolumn{4}{c|}{Macro-F1 Score}        & \multicolumn{4}{c|}{Micro - F1 Score} \\ \cline{2-9} 
                        & H               & S      & O      & M      & H       & S       & O       & M       \\ \hline
SE-ResNet18             & \textit{0.5139} & 0.4954 & 0.4863 & 0.4943 & 0.6895  & 0.6932  & 0.5101  & 0.5138  \\ \hline
ResNet18                & 0.4327          & 0.5033 & 0.5082 & 0.4303 & 0.7630  & 0.7838  & 0.5543  & 0.6229  \\ \hline
ResNet34                & 0.5226          & 0.5118 & 0.4921 & 0.5983 & 0.6645  & 0.6692  & 0.u5160  & 0.5553  \\ \hline
VGG                     & 0.5229          & 0.5158 & 0.4960 & 0.4964 & 0.6661  & 0.6837  & 0.5181  & 0.5607  \\ \hline
\end{tabular}
\caption{Task B - Class wise performance of our model with different backbones for visual feature extraction}
\label{res4}
\end{table*}
% Please add the following required packages to your document preamble:
% \usepackage{multirow}
\begin{table*}[!htb]
\centering
\captionsetup{justification=centering}
\begin{tabular}{|c|c|c|c|c|c|c|c|c|}
\hline
\multirow{2}{*}{Method} & \multicolumn{4}{c|}{Macro-F1 Score}        & \multicolumn{4}{c|}{Micro - F1 Score}      \\ \cline{2-9} 
                        & H               & S      & O      & M      & H      & S      & O               & M      \\ \hline
SE-ResNet18             & \textit{0.2650} & 0.2450 & 0.2459 & 0.4943 & 0.3253 & 0.3930 & \textit{0.3376} & 0.5138 \\ \hline
ResNet18                & 0.1317          & 0.2292 & 0.2114 & 0.4304 & 0.3243 & 0.4420 & 0.3819          & 0.6229 \\ \hline
ResNet34                & 0.2739          & 0.2669 & 0.2372 & 0.5183 & 0.3232 & 0.3732 & 0.3477          & 0.5554 \\ \hline
VGG                     & 0.2728          & 0.2686 & 0.2392 & 0.4964 & 0.3243 & 0.3597 & 0.3520          & 0.5607 \\ \hline
\end{tabular}
\caption{Task C - Class wise performance of our model with different backbones for visual feature extraction.}
\label{res5}
\end{table*}

\subsection{Feature Aggregation}
In this module, both image \& text features from their respective modules are combined to create a aggregated feature vector for multi-label hierarchical meme classification. Here, image feature vector ($F_{image}$) and text feature vector ($F_{text}$) are concatenated to get a single combined feature vector ($F_{combined}$) and then is fed into a dense layer to get prediction of overall sentiment class ($T_1$) of the meme image. As we know there is a strong correlation between overall sentiment class and the other semantic classes (because positive sentiment memes tends to have funny or motivational emotional expression and negative sentiment memes tends to have offensive or sarcasm emotional expression), prediction vector ($T_1$) was concatenated with combined feature vector and fed into a dense layer to get compressed memotion feature vector ($F_{memotion}$) for further fine-grained classification. This memotion feature vector is used with corresponding dense layer to get prediction vector for humour ($T_2$), sarcasm($T_3$), offensive($T_4$) \& motivational($T_5$) semantic Classes.   
\begin{equation}
    F_{combined} = F_{image} \otimes F_{text} 
\end{equation}
\begin{equation}
    F_{memotion} = f^{relu}_d(T_1 \otimes F_{combined})
\end{equation}
Here, $\otimes$ is denoted as concatenation operation and $f^{relu}_d$ is a mapping function for fully connected neural network or dense layer with relu activation function.
\[ T_1 = f^\sigma_{d_1}(F_{combined}) \]
\[ T_2 = f^\sigma_{d_2}(F_{memotion}) \]
\[ T_3 = f^\sigma_{d_3}(F_{memotion}) \]
\[ T_4 = f^\sigma_{d_4}(F_{memotion}) \]
\[ T_5 = f^\sigma_{d_5}(F_{memotion}) \]
Here, $f^{\sigma}_{d_i}$ is a mapping function for fully connected neural network or dense layer with sigmoid activation function for target $T_i$.

\section{Experiments \& Results}
We developed our models using Keras with Tensorflow as backend. The models were trained on a system with AMD Ryzen 1600X CPU, 16GB RAM and 8GB NVIDIA GTX 1080 GPU.  We  used  Adam  optimizer  to  train our network with batch size set as 16. Initial learning rate was set to 1e-4 which was decreasing exponentially per epoch. We trained our models until convergence. For calculating loss, we used binary cross-entropy. The results of Task A, B and C are shown in tables \ref{res1}, \ref{res2} and \ref{res3} respectively. The fine-grain class wise metric values of Task B and C are shown in tables \ref{res4} and \ref{res5} respectively. The values in italics are the officual best scores while the bold ones denote the results of a system that we had sent at a later stage to the organizers for evaluation, which used SE-ResNet18 as the visual backbone.

%\section{Discussion}

\section{Conclusion \& Future Work}
In the present work, we have build a system which uses multi-task learning to learn all the three tasks from SemEval 2020 Task 8 - memotional analysis as a single unified problem. The best macro F1-scores on Tasks A, B and C are 0.3488, 0.5112 and 0.3240 respectively. While for first task, SE-ResNet18 visual backbone gave the best results, for Task B \& C, ResNet34 gave the best result.In case of Task C, model performance performed worst because of imbalance data \& high intraclass correlation among fine grained class labels. In the future, we would like to work on topic modelling of social images into our pipeline to incorporate the domain knowledge.

\bibliography{semeval2018}
\bibliographystyle{acl_natbib}
\end{document}